\documentclass[5p, twocolumn]{elsarticle}
\pdfoutput=1 
\newcommand{\Tref}[1]{Table~\ref{#1}}
\newcommand{\eref}[1]{Eq.~(\ref{#1})}
\newcommand{\Eref}[1]{Equation~(\ref{#1})}
\newcommand{\fref}[1]{Fig.~\ref{#1}}
\newcommand{\Fref}[1]{Figure~\ref{#1}}

\newcommand{\Sref}[1]{Section~\ref{#1}}

\usepackage{color, times, graphicx, amsmath, multirow, subfigure, tabularx, caption, booktabs, hyperref}

\journal{Pattern Recognition}

\begin{document}

\begin{frontmatter}

\title{Approaching the Computational Color Constancy as a Classification Problem through Deep Learning}

%
%
%

\author[ys]{Seoung~Wug~Oh}
\ead{sw.oh@yonsei.ac.kr}

\author[ys]{Seon~Joo~Kim}
\ead{seonjookim@yonsei.ac.kr}

\address[ys]{Department of Computer Science, Yonsei University, 50 Yonsei-ro, Seodaemun-gu, Seoul, Republic of Korea}

\begin{abstract}
Computational color constancy refers to the problem of computing the illuminant color so that the images of a scene under varying illumination can be normalized to an image under the canonical illumination. 
In this paper, we adopt a deep learning framework for the illumination estimation problem.
The proposed method works under the assumption of uniform illumination over the scene and aims for the accurate illuminant color computation.
Specifically, we trained the convolutional neural network to solve the problem by casting the color constancy problem as an illumination classification problem.
We designed the deep learning architecture so that the output of the network can be directly used for computing the color of the illumination. 
Experimental results show that our deep network is able to extract useful features for the illumination estimation and our method outperforms all previous color constancy methods on multiple test datasets.
\end{abstract}


\end{frontmatter}


\section{Introduction}
Color constancy is the ability of the human vision system to ensure that perceived color of a scene remains relatively constant under varying illumination. 
The goal of the computational color constancy research is to have the computer emulate this capability of the human vision system. 

An image of a scene is photographed as follows:
\begin{equation} \label{eq:image_formation}
\rho_{k}(x) = \int E(\lambda)S(\lambda,x)R_{k}(\lambda)d\lambda \quad  k \in{\{R,G,B\}},
\end{equation}
where $\rho_{k}(x)$ is the intensity of each channel at pixel location $x$, $E(\lambda)$ is the illuminant spectrum, $S(\lambda,x)$ is the surface reflectance at pixel location $x$, and $R_{k}(\lambda)$ is the camera spectral sensitivity for each channel.
In the computational color constancy, the objective is to compute the chromaticity of illumination $\rho^{E}_{k}$:
\begin{equation} \label{eq:illumination}
\rho^{E}_{k} = \int E(\lambda)R_{k}(\lambda)d\lambda \quad  k \in{\{R,G,B\}}.
\end{equation}
The difficulty of solving for $\rho^{E}_{k}$ in \Eref{eq:illumination} lies in the ill-posedness of the problem as there are infinite number of combinations of the illuminant color and the surface color that result in the same image value $\rho$. 

In this paper, we adopt a deep learning framework to solve the color constancy problem.  
The deep learning has shown to be very useful for discovering hidden representations in large data. 
Among many deep learning systems, the convolutional neural network(CNN) has recently gained huge popularity due to its remarkable success in  object classification~\cite{krizhevsky2012imagenet}. 
Today, the CNN has been successfully applied in various computer vision tasks, including object recognition~\cite{simonyan2014very}, object detection~\cite{girshick2014rich}, face verification~\cite{taigman2014deepface}, and semantic segmentation~\cite{long2014fully}.

One of the principal factors behind the success of CNN on such a wide range of problems is that it does not require manually designed features for specific tasks. 
Instead, the system learns to extract useful features for a given task from a large number of training samples. 

Inspired by this learning capability of the CNN, we trained a deep learning architecture for our own task: the illumination estimation. 
We propose a deep learning based color constancy algorithm by casting the illumination estimation problem as a classification problem. 
By finding effective ways to transform the illumination estimation as an illumination classification problem and then to compute the accurate illumination chromaticity from the classification results, we are able to fully exploit the discriminating power of the CNN. 

Experimental results show that the CNN was able to extract useful features for the illumination estimation and our method outperforms all previous color constancy methods on multiple test datasets. 
While the CNN has shown to work very well for a variety of high level computer vision problems such as object recognition, the significance of this work lies in the fact that we are able to show that the CNN can be powerful in solving low level computer vision problems (color or appearance based) as well.

\section{Related Work}
Computing the illumination from a given image is an active research topic in computer vision and the work on this problem can be divided into two categories: statistics-based methods and learning-based methods.

\subsection{\bf Statistics-based Illumination Estimation}
Statistics-based methods estimate the illumination with a strong assumption on the scene statistics. 
One major line of statistic-based method is based on assumptions on the greyness of the scene color statistics.
The well-known Grey-World~\cite{buchsbaum1980spatial} and White-Patch~\cite{funt2010rehabilitation,land1971lightness} algorithms and their extended versions (Shades of Grey~\cite{finlayson2004shades} and Grey-Edge~\cite{van2007edge}) fall into this category. 
The methods in this group assume certain kinds of reflectance statistics (e.g., average reflectance, max reflectance, and average reflectance difference) in the scene to be achromatic.

Another line of statistics-based methods estimate the illumination by analyzing the physical property of the scene. They exploit the statistics of the bright and the dark pixels~\cite{cheng2014illuminant,joze2012role}, specular highlights~\cite{drew2012specularity, lee1986method}, or grey pixels~\cite{yang2015efficient} as important cues for the illumination estimation.

Recently, understanding the mechanism of the human visual system (HVS) has been found to be useful in building the statistical assumptions by mimicking the human built-in ability of color constancy~\cite{gao2014efficient,gao2015color}. One limitation of these approaches is that we are still far from fully understanding the mechanism of the HVS. In this regard, we believe our deep learning approach which train HVS inspired complex model (CNN) with large data to simulate color information processing on the human brain could make a breakthrough for the problem of the computational color constancy because we do not need to know how it works specifically.

Although the statistics-based methods are computationally efficient and do not require training data, the performance of these methods are usually not on par with the learning-based methods. 

\subsection{\bf Learning-based Illumination Estimation}
Learning-based methods can be further categorized into two groups according to what they learn: combinatorial methods and direct methods. 

Combinatorial methods find the best combination of the statistics-based methods for an input image based on the scene contents. Various scene characteristics, including scene semantics~\cite{van2007using}, indoor/outdoor classification~\cite{bianco2008improving}, 3D geometry~\cite{lu2009color}, low-level visual properties~\cite{bianco2010automatic}, and natural image statistics~\cite{gijsenij2011color} are used to find the best combination. Refer to the survey paper~\cite{li2014evaluating} for more information.

Direct methods build their own estimation model and estimate the illumination by learning the model from the training data.
Gamut-based methods~\cite{barnard2000improvements,forsyth1990novel,gijsenij2010generalized} find the canonical gamut from the training data and estimate the illumination by mapping the gamut of the input image into the canonical gamut.
Distributions of the pixel intensity and the chromaticity are used as the key features for estimating the illumination in 
the correlation framework~\cite{finlayson2001color}, the neural network based method~\cite{cardei2002estimating}, the support vector regression~\cite{xiong2006estimating}, and the Bayesian framework~\cite{gehler2008bayesian,rosenberg2003bayesian}.
In~\cite{chakrabarti2012color,gijsenij2010generalized}, the derivative structure and the spatial distribution of the image are used for the illumination estimation.

Recent studies show that relatively simple features related to the color statistics can be used to give accurate results with computationally efficient machine learning techniques~\cite{cheng2015effective,finlayson2013corrected, chakrabarti2015color}.
It is also shown that the high and the mid-level representation of the scene, in addition to the chromatic features, are useful for data driven approaches. 
Both the surface texture feature and the color histogram are used for the exemplar-based learning that finds similar surfaces~\cite{joze2014exemplar} and for the optimization of bilayer sparse coding model for the illumination estimation~\cite{li2013illumination}.

The selection of features is one of the most important part in the learning based methods. 
In most learning based work, the features are manually selected based on heuristics and simple assumptions.
Recently, deep learning (CNN) based systems for the illumination estimation has been proposed~\cite{bianco2015color-constancy,LouBMVC2015}. 
These methods integrate the feature learning and the regression by minimizing the Euclidean loss on the illuminant color.
They show that relatively shallow network~\cite{bianco2015color-constancy} or deep network~\cite{LouBMVC2015} trained to regress input image into the illuminant chromaticity can produce good results.

The biggest difference between our framework and previous learning based methods is that we transform the illumination estimation problem to a classification problem on real illuminants. In previous learning based work, the illumination estimation is often considered as a regression based on various features. In our deep learning based framework, the features that are useful for distinguishing the training samples under different illuminants are learned automatically. We show that discriminative learning based on the CNN outperform previous learning-based color constancy algorithms.

\begin{figure*}
\centering
\subfigure{\includegraphics[width=1.0\linewidth]{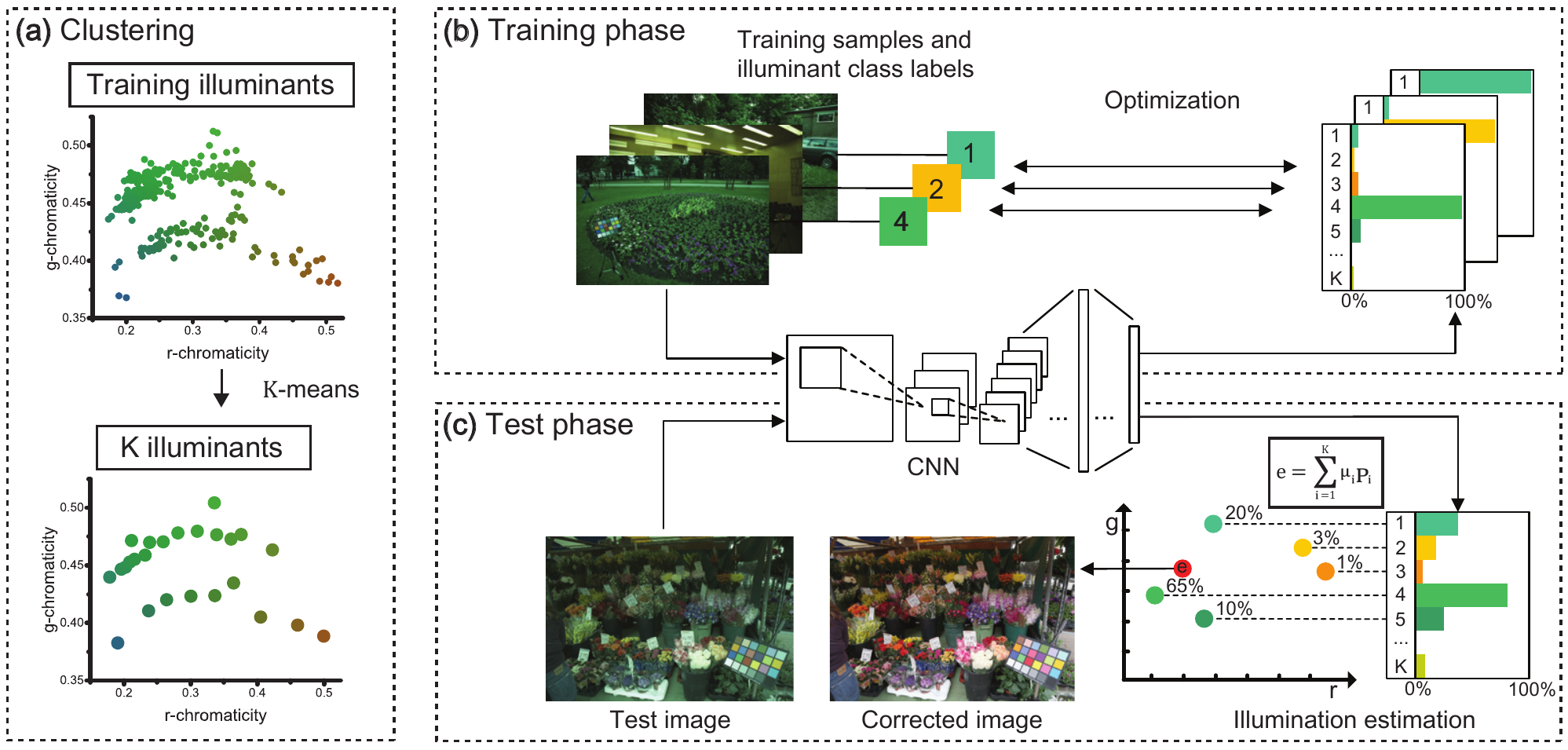}}  
\caption{The overview of our system for the computational color constancy.} 
\label{fig:overview}
\end{figure*}

\section{Convolutional Neural Network for Illumination Estimation}

\subsection{System Overview}
\Fref{fig:overview} shows the overview of our system. 
The training images are first clustered according to the illumination color assigned to each image (\fref{fig:overview}(a)) and the images with the new labels are used to train our convolutional neural network system. 
Our deep network is designed to output the probabilities of the given image belonging to each illumination cluster (\fref{fig:overview}(b)). 
After the training, the illumination color of a test image is computed by combining the outputs from our CNN  as shown in \fref{fig:overview}(c).

\begin{figure} 
\centering
\subfigure{\includegraphics[width=0.90\linewidth]{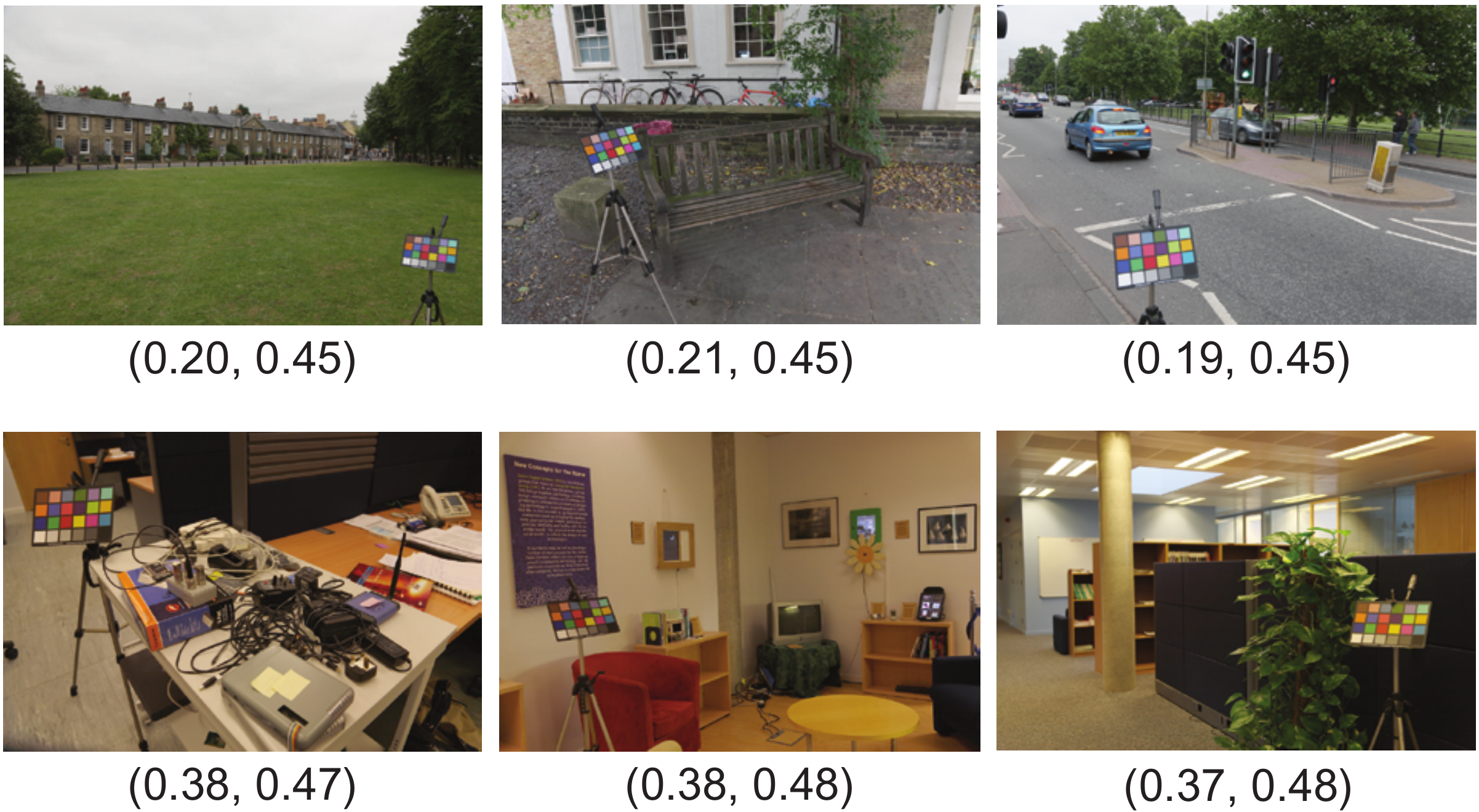}}
\caption{Some examples of images under similar lighting conditions in the Gehler-Shi dataset \cite{Shi_cc}. The ground truth illuminant in $rg$ chromaticity space is shown under each image. We cannot use the given illuminant directly as the labels to train the CNN because the illuminations are too similar. The classification becomes easier when clustering similar illuminations (each row of images in this example) into a new label.}
\label{fig:cluster}
\vspace{-0.4cm}
\end{figure}

\subsection{Clustering the Illuminants}
One of the key factors in enabling the CNN to work well with the illumination estimation is the illumination clustering stage.
In principal, the CNN learns multi-scale features that can discriminate different classes the best.
Therefore, one can expect the CNN to work well for the problems in which the classes are well separable and not so well for the cases where the classes are not so well distinguished.
The illumination estimation problem falls into the latter case without further groupings, because there exist many similar illuminations in the world. 
For example, while 568 images were given in the Gehler-Shi dataset \cite{Shi_cc}, many of the images are taken under very similar lighting conditions as shown in \fref{fig:cluster}. 
If we directly use this type of training data with many similar illuminations, it is a little too much to expect the CNN to learn to discriminate all those similar illuminations well. 
Therefore, we propose to first cluster the illuminations in the training data to group similar illuminations so that the classes become further apart, which makes the classification easier for the CNN.   

For the clustering, we use the K-means algorithm with the angular distance $\delta_{angle}$ as the distance measure:
\begin{equation}
\delta_{angle}(e_{a}, e_{b}) = \cos^{-1}\Big({\frac{\lVert e_{a} \cdotp e_{b} \rVert}{ \lVert e_{a} \rVert \lVert e_{b} \rVert}}\Big),
\end{equation}
where $e_{a}$ and $e_{b}$ are the given illuminant color for images $a$ and $b$. 
We empirically chose $K = 25$ for the Gehler-Shi set~\cite{Shi_cc}, $K=20$ for Gray-Ball set~\cite{ciurea2003large}, and $K=50$ for NUS 8-camera set~\cite{cheng2014illuminant}. The effect of $K$ is discussed in \Sref{discussion}.
After the clustering, the training images with the new illumination labels are fed into our CNN for the learning.

\subsection{Convolutional Neural Network Architecture}
\begin{figure}
\noindent\begin{minipage}{\linewidth}
\centering
\includegraphics[width=1\linewidth]{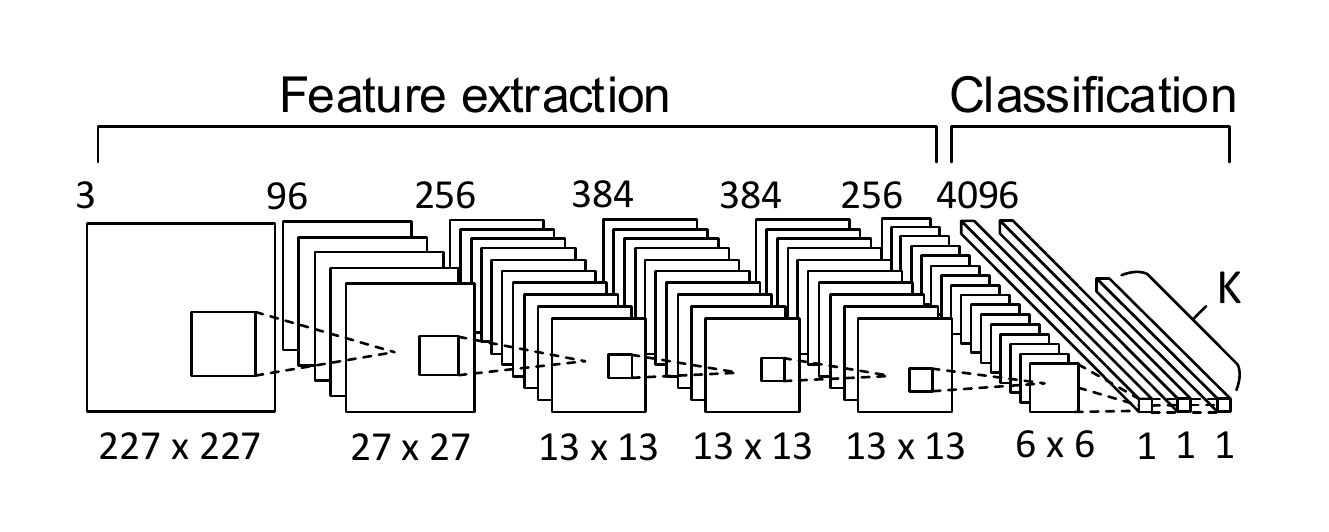}
\vspace{-6mm}
\captionof{figure}{The structure of our deep convolutional neural network.}
\label{fig:cnn}

\captionof{table}{The parameters of our deep convolutional neural network. Layer type Conv. and FC refer to convolution and fully-connected layer, respectively.}
\footnotesize
\begin{tabular}{ c c c c c c c c }
Type & Kernel & Stride & Pad & Outputs & Activ. & Regular.\\
\specialrule{.15em}{.1em}{.1em} 
Conv. & 11 & 4 & 0 & 96 & ReLU &  - \\
Max pool & 3 & 2 & 0 & 96 & - &  - \\
LRN & - & - & - & 96 & - & - \\
\hline
Conv. & 5 & 1 & 2 &256 & ReLU &  - \\
Max pool & 3 & 2 & 0 & 256 & - &  - \\
LRN & - & - & - & 256 & - & - \\
\hline
Conv. & 3 & 1 & 1 & 384 & ReLU &  - \\
\hline
Conv. & 3 & 1 & 1 & 384 & ReLU & - \\
\hline
Conv. & 3 & 1 & 1 & 256 & ReLU & - \\
\hline
FC & - & - & - & 4096 & ReLU & Dropout \\
\hline
FC & - & - & - & 4096 & ReLU & Dropout \\
\hline
FC & - & - & - & K & Softmax & -\\
\hline
\end{tabular}
\label{table:CNN_table}
\end{minipage} 
\end{figure}

We adopt the CNN architecture proposed in ~\cite{krizhevsky2012imagenet}, which has shown to work well for various problems~\cite{donahue2013decaf,eigen2014depth}.
From this basic structure, we have made the following changes in the network. 
The sparse connections in the layers 3,4,5 for multi-GPU implementation are replaced with dense connections for the single GPU system. 
The number of output units in the final fully-connected layer is of course changed to match the number of classes $K$ used in our problem. 
As illustrated in \fref{fig:cnn}, the CNN consists of five feature extraction layers of the convolution and the max-pooling, followed by three fully-connected layers. 
Other hyper-parameters of the network including the filter size, the strides, and the number of feature map are shown in \fref{fig:cnn}. 
The rectified linear unit (ReLU) activation is applied to the response of every hidden layer except for the final fully-connected layer, which is directly connected to the softmax layer. 
The responses of the first and the second feature extraction layers are normalized as done in~\cite{krizhevsky2012imagenet} and the dropout~\cite{hinton2012improving} is applied to the fully-connected layer 6 and 7 for the regularization.
As the CNN is used for the illumination classification, we place the softmax layer that is often used to represent the probability distribution over classes at the end of network~\cite{krizhevsky2012imagenet, bridle1990probabilistic, Goodfellow-et-al-2016-Book}.
Therefore, our CNN outputs a vector of length $K$, of which the elements are always positive and sum to 1. 
The ultimate goal of our system is to compute the chromaticity of the illumination, so we must infer the final color from the probability.
The illumination estimation procedure is described in \Sref{sect:illumination_estimation}.

%

\subsection{Training strategy}
One well known nature of the deep learning is that it requires a large amount of data to train the deep structures to extract the generalized feature representation and to avoid the overfitting.
This is a problem for using the deep learning for color constancy because the available datasets for color constancy do not contain sufficient data. 
The Gehler-Shi set provides 568 images, the Gray-Ball set provides 11,346 images (highly correlated as they are from video clips), and the NUS 8-camera set provides 1,736 images from 8 different cameras resulting in about 200 distinct scenes. 
To deal with the lack of data, the network is first trained with a very large dataset used for the object classification and then the learned weights are used as the initial weights to optimize the network for our own task.
This type of strategy is called the transfer learning~\cite{oquab2014learning} and we pretrained our deep network using the ImageNet database ~\cite{deng2009imagenet} which contains 1.2 million images.
With this strategy, the optimization converges much faster than starting from the scratch and it also prevents the network from overfitting to a small amount of the training data. 

To further overcome the lack of training data and to prevent overfitting, we also augmented the training data by adding more data through various transformations:
\begin{itemize}\itemsep-2pt
	\item Rotation: Input images are rotated by $r \in [-10,-5,0,5,10]$ degrees.
	\item Translation: Image patches are cropped at random image positions. 
  \item Scale: Input images are scaled while we resize cropped patches with different size (from 250 to 1000) to CNN input size.
	\item Flips: Input images are horizontally flipped.
\end{itemize}
We make 200 input patches for CNN from an image. In the test phase, the same transformations are applied to test images making 200 input patches and the results of CNN from each patch is averaged.

It should be emphasized that the transforms that alter the RGB color values may lead to the loss of essential cues for the illuminant classification, so we only augment the training samples with color-preserving transformations that do not affect the chromatic information of the images.

The optimization procedure of our CNN for the illumination classification is similar to the one used for the object classification~\cite{krizhevsky2012imagenet}.
The network is optimized to output probability distributions over possible illuminant classes. 
For a given input image $\bf{x}$, the CNN $\mathcal{F}$ is learned to predict the probability distribution over $K$ illuminant classes $\bf{\hat{y}} \in [0,1]^K$: 
\begin{equation}
{\bf \hat{y}} = \mathcal{F}(\bf{x}),
\end{equation}
where the $k^{th}$ element of ${\bf \hat{y}}$ represents $P(y = k | {\bf x})$.
The weights of the CNN are learned by minimizing the following multinomial negative log-likelihood loss $L$:
\begin{equation}
L({\bf x}, c) = -\log(\hat{\bf{y}}_c) = -\log(P(y = c | x,W)),
\label{eq:loss}
\end{equation}
where ${\bf \hat{y}} = \mathcal{F}(\bf{x})$ is the output of the CNN, $\hat{\bf{y}}_c$ is the $c^{th}$ element of $\hat{\bf{y}}$, $c$ is the ground truth label for image {\bf x}, and the $W$ is the weight of the network.

The network parameters are updated by the stochastic gradient descent with the batch size of 100 samples, the momentum of 0.9, and the weight decay of 0.0005. 
The derivative of the loss is back-propagated through the network to update the entire network parameters.
We set the learning rate differently by the layers: the learning rate for first convolution layer and the last two fully-connected layers are 10 times larger than the other layers.
This is because the first and the last layers should be more tuned for our specific task as the system is first pretrained for different data and task (object classification).

\begin{figure}
\centering
\subfigure{\includegraphics[width=1.0\linewidth]{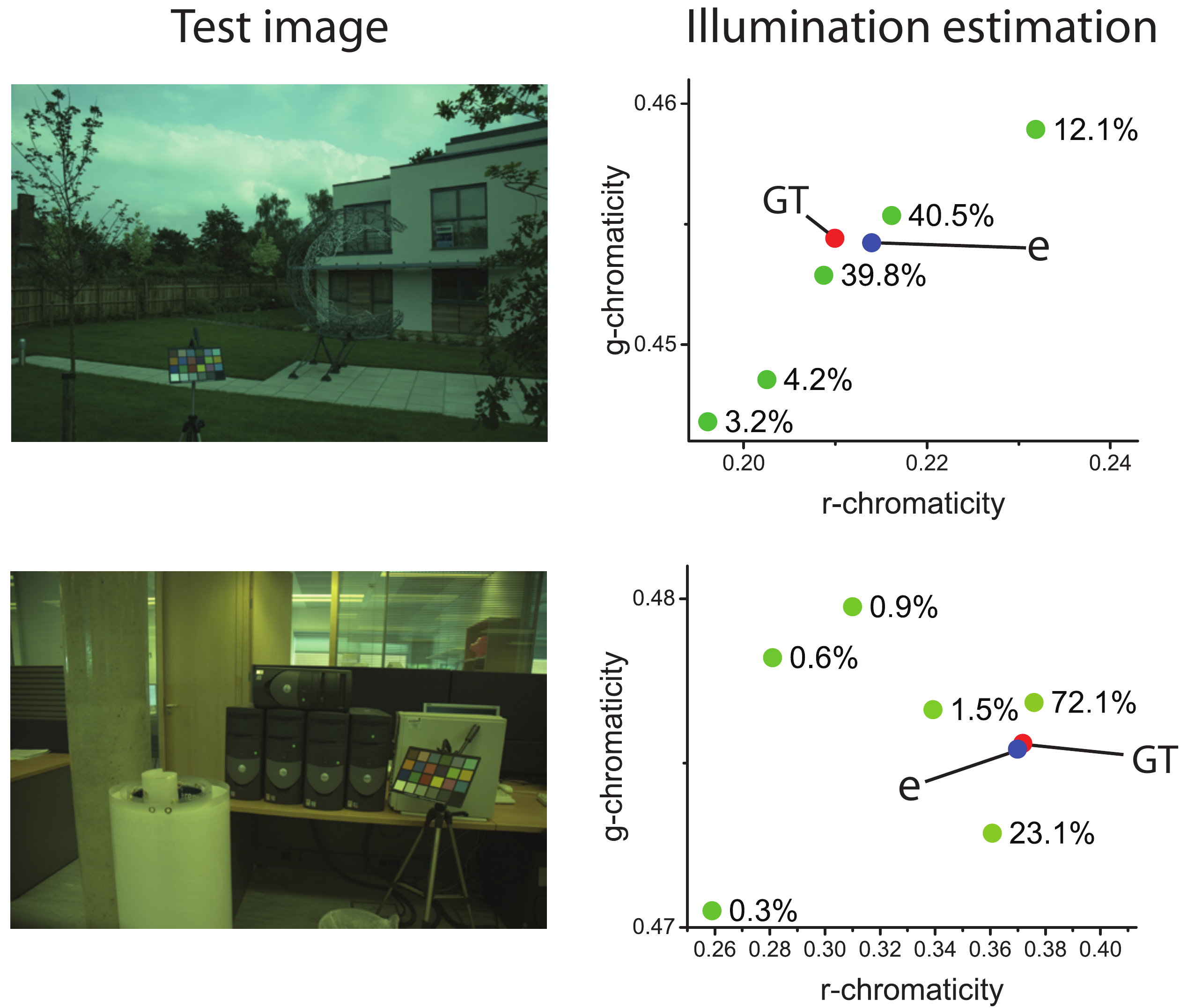}}
\caption{Computing the illuminant chromaticity $e$ from the output of the CNN. The green dots represent the cluster centers and the percentage numbers indicate the probability of the given image belonging to that class. The illuminant is computed as a weighted sum of the cluster centers with the weights being the probabilities. The ground truth are shown as red dots with the label GT.}
\label{fig:real_ex}
\end{figure}

\subsection{Illumination Estimation} \label{sect:illumination_estimation}
Given the network $\mathcal{F}$ which is trained according to the loss function~\eref{eq:loss}, we estimate the single illumination color of a test image from the prediction: ${\bf \hat{y}} = \mathcal{F}(\bf{x})$.
As explained, our CNN $\mathcal{F}$ is trained to predict the probability of a given image $\bf{x}$ belonging to one of $K$ illumination classes \cite{bridle1990probabilistic,Goodfellow-et-al-2016-Book}: 
\begin{equation} \label{eq:output}
{\bf \hat{y}} = 
\begin{pmatrix}
  P(y = 1 | x)\\
  P(y = 2 | x)\\
  \vdots\\
  P(y = K | x) 
 \end{pmatrix} 
\end{equation}

Since our goal is to compute the illumination of a given image not just the cluster index among $K$ groups, we need to compute the final illuminant chromaticity from the output in \Eref{eq:output}. One trivial solution is to take the center of the cluster that gives maximum probability ($\arg\,\min_k\,{\bf \hat{y}}_k$). However, this approach is not suitable for estimating the illuminant color because it limits the possible illuminant to $K$ illuminants.
For this purpose, we estimate the final illuminant chromaticity $e$ by computing the weighted average of the cluster centers ($\mu_i$) with the weights being $P(y=i|x)$ from the CNN output:
\begin{equation}
 e = \sum\limits_{i=1}^K \mu_{i} {\bf \hat{y}}_{i} .
\end{equation} 
With this approach, our system can estimate illuminants in a more flexible way as seen in ~\fref{fig:real_ex}.


\begin{table*}[ht!] \setlength{\tabcolsep}{1mm}
\caption{Performance of various methods on the Gehler-Shi set. Most results of previous methods are directly from~\cite{cccom, gijsenij2011computational}, and results of Bilayer Sparse-Coding, 19 Edge Moments, and Simple Feature Regression are from~\cite{li2013illumination},~\cite{finlayson2013corrected}, and ~\cite{cheng2015effective}, respectively.}
\label{table:CC_result}
\centering 
\begin{tabular}{|c|l|c|c|c|c|c|}
\hline
\multicolumn{2}{|c|}{Method} & Mean & Median & Trimean & Best-25\% & Worst-25\% \\
\hline
\hline
\multirow{9}{*}{Statistics-Based} 
&Grey-world~\cite{buchsbaum1980spatial} & 6.36 & 6.28 & 6.28 & 2.33 & 10.58 \\
&White-patch~\cite{land1971lightness} & 7.55 & 5.68 & 6.36 & 1.45 & 16.15 \\
&Shades-of-Grey~\cite{finlayson2004shades} & 4.93 & 4.01 & 4.23 & 1.14 & 10.22  \\
&General Grey-world~\cite{van2007edge} & 4.67 & 3.46 & 3.81 & 0.98 & 10.21  \\
&$1^{st}$-order Grey-Edge~\cite{van2007edge} & 5.33 & 4.52 & 4.73 & 1.86 & 10.05 \\
&$2^{nd}$-order Grey-Edge~\cite{van2007edge} & 5.13 & 4.44 & 4.63 & 2.11 & 9.27 \\
&Bright-and-dark Colors PCA~\cite{cheng2014illuminant} & 3.52 & 2.14 & 2.47 & 0.50 & 8.74  \\
&Double-Opponency~\cite{gao2015color} & 3.98 & 2.43 & - & - & 9.08 \\
&Local Surface Reflectance~\cite{gao2014efficient} & 3.31 & 2.80 & 2.87 & 1.14 & 6.39  \\
&Grey Pixels~\cite{yang2015efficient} & 4.60 & 3.10 & - & - & -  \\
\hline
\hline
\multirow{18}{*}{Learning-Based} 
&Pixel-based Gamut~\cite{gijsenij2010generalized}  & 4.20 & 2.33 & 2.92 & 0.50 & 10.75  \\
&Edge-based Gamut~\cite{gijsenij2010generalized} & 6.72 & 5.60 & 5.80 & 2.05 & 13.50 \\
&Intersection-based Gamut~\cite{gijsenij2010generalized}  & 4.21 & 2.34 & 2.91 & 0.50 & 10.78 \\
&Regression (SVR)~\cite{xiong2006estimating} & 8.08 & 6.73 & 7.19 & 3.35 & 14.92  \\
&Bayesian~\cite{gehler2008bayesian} & 4.82 & 3.46 & 3.89 & 1.26 & 10.51  \\
&Spatio-spectral~\cite{chakrabarti2012color} & 3.59 & 2.96 & 3.05 & 0.91 & 7.45  \\
&High-level Visual Information~\cite{van2007using} & 3.48 & 2.47 & 2.61 & 0.84 & 8.03  \\
&Natural Image Statistics~\cite{gijsenij2011color} & 4.19 & 3.13 & 3.45 & 1.00 & 9.24  \\
&CART-based Combination~\cite{bianco2010automatic} & 3.90 & 2.91 & 3.23 & 1.02 & 8.29  \\
&Bilayer Sparse-Coding~\cite{li2013illumination} & 4.00 & 2.50 & 2.80 & 1.00 & 10.80  \\
&Exemplar-based~\cite{joze2014exemplar} & 2.89 & 2.27 & 2.42 & 0.82 & 5.98  \\
&19 Edge Moments~\cite{finlayson2013corrected} & 2.80 & 2.00 & - & - & -  \\
&Simple Feature Regression~\cite{cheng2015effective} & 2.42 & 1.65 & 1.75 & 0.38 & 5.87  \\
&Alexnet+SVR~\cite{bianco2015color-constancy} & 4.74 & 3.09 & 3.52 & 1.10 & 11.11  \\
&CNN Regression~\cite{bianco2015color-constancy} & 2.63 & 1.98 & 2.13 & 0.74 & 5.64 \\
&Luminance-to-Chromaticity~\cite{chakrabarti2015color} & 2.56 & 1.67 & 1.89 & - & - \\
&NetColorChecker~\cite{LouBMVC2015} & 3.10 & 2.30 & - & - & - \\
&{\bf Proposed} & {\bf 2.16} &{\bf 1.47} &{\bf 1.61} &{\bf 0.37} &{\bf 5.12} \\
\hline
 \end{tabular}
\vspace{6mm}
\end{table*}

\begin{table*}[ht!]  \setlength{\tabcolsep}{1mm}
\caption{Performance of various methods on the SFU Gray-ball set (linear). All the methods use linearized images and errors reported here are based on the recomputed ground truth~\cite{gijsenij2011computational}. All the learning-based methods use the 15-fold cross-validation.}
\label{table:gb_result}
\centering 
\begin{tabular}{|c|l|c|c|c|c|c|}
\hline
\multicolumn{2}{|c|}{Method} & Mean & Median & Trimean & Best-25\% & Worst-25\%  \\
\hline
\hline
\multirow{6}{*}{Statistics-Based} 
&Grey-world~\cite{buchsbaum1980spatial} & 13.01 & 10.96 & 11.53 & 3.15 & 25.95  \\
&White-patch~\cite{land1971lightness} & 12.68 & 10.50 & 11.25 & 2.52 & 26.19  \\
&Shades-of-Grey~\cite{finlayson2004shades} & 11.55 & 9.70 & 10.23 & 3.36 & 22.72  \\
&General Grey-world~\cite{van2007edge} & 11.55 & 9.70 & 10.23 & 3.36 & 22.72  \\
&$1^{st}$-order Grey-Edge~\cite{van2007edge} & 10.58 & 8.84 & 9.18 & 3.01 & 21.14  \\
&$2^{nd}$-order Grey-Edge~\cite{van2007edge} & 10.68 & 9.02 & 9.40 & 3.22 & 20.89 \\
\hline
\hline
\multirow{10}{*}{Learning-Based}
&Pixel-based Gamut~\cite{gijsenij2010generalized}  & 11.79 & 8.88 & 9.97 & 2.78 & 24.94  \\
&Edge-based Gamut~\cite{gijsenij2010generalized} & 12.78 & 10.88 & 11.38 & 3.56 & 25.04  \\
&Intersection-based Gamut~\cite{gijsenij2010generalized}  & 11.81 & 8.93 & 10.00 & 2.80 & 24.94 \\
&Regression (SVR)~\cite{xiong2006estimating} & 13.14 & 11.24 & 11.75 & 4.42 & 25.02  \\
&Spatio-spectral~\cite{chakrabarti2012color} & 10.31 & 8.89 & 9.16 & 2.80 & 20.31  \\
&High-level Visual Information~\cite{van2007using} & 9.73 & 7.71 & 8.17 & 2.33 & 20.59  \\
&Natural Image Statistics~\cite{gijsenij2011color} & 9.87 & 7.65 & 8.29 & 2.42 & 20.84  \\
&Bilayer Sparse-Coding~\cite{li2013illumination} & 9.20 & 7.30 & 7.80 & 2.10 & 19.60  \\
&Exemplar-based~\cite{joze2014exemplar} & 7.97 & 6.46 & 6.77 & 2.01 &  16.61  \\
&{\bf Proposed} & {\bf 6.60} &{\bf 4.19} &{\bf 4.72} &{\bf 1.27} & {\bf 16.09}  \\
\hline
 \end{tabular}
\vspace{1mm}
\end{table*}

\renewcommand{\arraystretch}{0.85}
\begin{table*} \setlength{\tabcolsep}{1mm}
\caption{Performance of various methods on NUS 8-camera set. Methods reported on this table are Grey-world (GW),
White-patch (WP), Shades-of-grey (SG), General Grey-world (GGW), $1^{st}$-order Grey-edge (GE1), $2^{nd}$-order Greyedge
(GE2), Bright-and-dark Colors PCA (PCA), Local Surface Reflectance Statistics (LSR), Pixels-based Gamut (PG),
Edge-based Gamut (EG), Bayesian framework (BF), Spatio-spectral Statistics (SS), Natural Image Statistics (NIS),
Corrected-moment method (CM), and Our method. Results of previous methods are directly from~\cite{cheng2014illuminant,cheng2015effective}.}
\label{table:nus_full_result}
\centering 
\vspace{2pt}
\begin{tabular}{|c|c|c|c|c|c|c|c|c|c|c|c|c|c|c|c|c|}
\hline
\multicolumn{1}{|c|}{\multirow{2}{*}{Method}} & GW & WP & SG & GGW & GE1 & GE2 & PCA & LSR & PG & EG & BF & SS & NIS & CM & SF & \multirow{2}{*}{Ours} \\
& \cite{buchsbaum1980spatial} & \cite{land1971lightness} & \cite{finlayson2004shades} & \cite{van2007edge} & \cite{van2007edge} & \cite{van2007edge} & \cite{cheng2014illuminant} & \cite{gao2014efficient} & \cite{gijsenij2010generalized} & \cite{gijsenij2010generalized} & \cite{gehler2008bayesian} & \cite{chakrabarti2012color} & \cite{gijsenij2011color} & \cite{finlayson2013corrected} & \cite{cheng2015effective} & \\

\hline
\hline
\multicolumn{17}{|c|}{Canon EOS-1Ds Mark III} \\
\hline
Mean& 5.16 & 7.99 & 3.81 & 3.16 & 3.45 & 3.47 & 2.93 & 3.43 & 6.13 & 6.07 & 3.58 & 3.21 & 4.18 & 2.94 & {\bf 2.26} & 2.57 \\
Median& 4.15 & 6.19 & 2.73 & 2.35 & 2.48 & 2.44 & 2.01 & 2.51 & 4.30 & 4.68 & 2.80 & 2.67 & 3.04 & 1.98 & {\bf 1.57} & 2.18 \\
Worst-25\%& 11.00 & 16.75 & 8.52 & 7.08 & 7.69 & 7.76 & 6.82 & 7.30 & 14.16 & 13.35 & 7.95 & 6.43 & 9.51 & 6.93 & {\bf 5.17} & 4.76 \\
\hline
\hline
\multicolumn{17}{|c|}{Canon EOS 600D} \\ 
\hline
Mean& 3.89 & 10.96 & 3.23 & 3.24 & 3.22 & 3.21 & 2.81 & 3.59 & 14.51 & 15.36 & 3.29 & 2.67 & 3.43 & 2.76 & 2.43 & {\bf 1.85} \\
Median& 2.88 & 12.44 & 2.58 & 2.28 & 2.07 & 2.29 & 1.89 & 2.72 & 14.83 & 15.92 & 2.35 & 2.03 & 2.46 & 1.85 & {\bf 1.62} & 1.75 \\
Worst-25\%& 8.53 & 18.75 & 7.06 & 7.58 & 7.48 & 7.41 & 6.50 & 7.40 & 18.45 & 18.66 & 7.93 & 5.77 & 7.76 & 6.28 & 5.63 & {\bf 3.00} \\
\hline
\hline
\multicolumn{17}{|c|}{Fujifilm X-M1} \\ 
\hline
Mean& 4.16 & 10.20 & 3.56 & 3.42 & 3.13 & 3.12 & 3.15 & 3.31 & 8.59 & 7.76 & 3.98 & 2.99 & 4.05 & 3.23 & {\bf 2.45} & 2.97 \\
Median& 3.30 & 10.59 & 2.81 & 2.60 & 1.99 & 2.00 & 2.15 & 2.48 & 8.87 & 8.02 & 3.20 & 2.45 & 2.96 & 2.11 & {\bf 1.58} & 2.75 \\
Worst-25\%& 9.04 & 18.26 & 7.55 & 7.62 & 7.33 & 7.23 & 7.30 & 7.06 & 13.40 & 13.44 & 8.82 & 5.99 & 9.37 & 7.66 & 5.73 & {\bf 5.20} \\
\hline
\hline
\multicolumn{17}{|c|}{Nikon D5200} \\ 
\hline
Mean& 4.38 & 11.64 & 3.45 & 3.26 & 3.37 & 3.47 & 2.90 & 3.68 & 10.14 & 13.00 & 3.97 & 3.15 & 4.10 & 3.46 & 2.51 & {\bf 2.25} \\
Median& 3.39 & 11.67 & 2.56 & 2.31 & 2.22 & 2.19 & 2.08 & 2.83 & 10.32 & 12.24 & 3.10 & 2.26 & 2.40 & 2.04 & {\bf 1.65} & 2.00 \\
Worst-25\%& 9.69 & 21.89 & 7.69 & 7.53 & 8.42 & 8.21 & 6.73 & 7.57 & 15.93 & 24.33 & 8.18 & 6.90 & 10.01 & 8.64 & 5.98 & {\bf 3.86} \\
\hline
\hline
\multicolumn{17}{|c|}{Olympus E-PL6} \\ 
\hline
Mean& 3.44 & 9.78 & 3.16 & 3.08 & 3.02 & 2.84 & 2.76 & 3.22 & 6.52 & 13.20 & 3.75 & 2.86 & 3.22 & 2.95 & {\bf 2.26} & 2.64 \\
Median& 2.58 & 9.50 & 2.42 & 2.15 & 2.11 & 2.18 & 1.87 & 2.49 & 4.39 & 8.55 & 2.81 & 2.21 & 2.17 & 1.84 & {\bf 1.52} & 2.22 \\
Worst-25\%& 7.41 & 18.58 & 6.78 & 6.69 & 6.88 & 6.47 & 6.31 & 6.55 & 15.42 & 30.21 & 8.19 & 6.14 & 7.46 & 7.39 & 5.38 & {\bf 5.14} \\
\hline
\hline
\multicolumn{17}{|c|}{Panasonic Lumix DMC-GX1} \\
\hline
Mean& 3.82 & 13.41 & 3.22 & 3.12 & 2.99 & 2.99 & 2.96 & 3.36 & 6.00 & 5.78 & 3.41 & 2.85 & 3.70 & 3.10 & 2.36 & {\bf 1.84} \\
Median& 3.06 & 18.00 & 2.30 & 2.23 & 2.16 & 2.04 & 2.02 & 2.48 & 4.74 & 4.85 & 2.41 & 2.22 & 2.28 & 1.77 & 1.61 & {\bf 1.53} \\
Worst-25\%& 8.45 & 20.40 & 7.12 & 6.86 & 7.03 & 6.86 & 6.66 & 7.42 & 12.19 & 11.38 & 8.00 & 5.90 & 8.74 & 7.81 & 5.65 & {\bf 3.37} \\
\hline
\hline
\multicolumn{17}{|c|}{Samsung NX2000} \\ 
\hline
Mean& 3.90 & 11.97 & 3.17 & 3.22 & 3.09 & 3.18 & 2.91 & 3.84 & 7.74 & 8.06 & 3.98 & 2.94 & 3.66 & 2.74 & 2.53 & {\bf 1.89} \\
Median& 3.00 & 12.99 & 2.33 & 2.57 & 2.23 & 2.32 & 2.03 & 2.90 & 7.91 & 6.12 & 3.00 & 2.29 & 2.77 & 1.85 & 1.78 & {\bf 1.65} \\
Worst-25\%& 8.51 & 20.23 & 6.92 & 6.85 & 7.00 & 7.24 & 6.48 & 7.98 & 13.01 & 16.27 & 8.62 & 6.22 & 8.16 & 6.27 & 5.96 & {\bf 3.44} \\
\hline
\hline
\multicolumn{17}{|c|}{Sony SLT-A57} \\ 
\hline
Mean& 4.59 & 9.91 & 3.67 & 3.20 & 3.35 & 3.36 & 2.93 & 3.45 & 5.27 & 4.40 & 3.50 & 3.06 & 3.45 & 2.95 & {\bf 2.15} & 3.25 \\
Median& 3.46 & 7.44 & 2.94 & 2.56 & 2.58 & 2.70 & 2.33 & 2.51 & 4.26 & 3.30 & 2.36 & 2.58 & 2.88 & 1.85 & {\bf 1.40} & 3.11 \\
Worst-25\%& 9.85 & 21.27 & 7.75 & 6.68 & 7.18 & 7.14 & 6.13 & 7.32 & 11.16 & 9.83 & 8.02 & 6.17 & 7.18 & 6.89 & {\bf 4.99} & 5.27 \\
\hline
 \end{tabular}
\end{table*}

\section{Experiments} \label{sect:experiments}
In this section, we evaluate our CNN based color constancy algorithm which is implemented by using the Caffe package~\cite{jia2014caffe} 
on a machine with a NVIDIA GTX 970 GPU.
We compare our method with many existing methods using the publicly available datasets and the following error metric was used for the comparisons:
\begin{equation}
err_{angle}(e_{a}, e_{b}) = \frac{180^{\circ}}{\pi}\cos^{-1}\Big({\frac{\lVert e_{a} \cdotp e_{b} \rVert}{ \lVert e_{a} \rVert \lVert e_{b} \rVert}}\Big),
\end{equation}
where $e_{a} \cdotp e_{b}$ is the dot product of $e_{a}$ and $e_{b}$, and  $\lVert \cdotp \rVert$ denotes the Euclidean norm. 
For fair comparisons, we follow the same experimental settings as in \cite{gijsenij2011computational} and use the cross validation strategy to test every image in the dataset.
For the error values of other methods, we directly use the values reported in previous works.

We first test our method using the Gehler-Shi set~\cite{gehler2008bayesian,Shi_cc} which contains 568 high quality images from two DSLR cameras. To evaluate the entire dataset, we use the 3-fold cross validation as done in \cite{gijsenij2011computational}. The color checker inside each image is masked out before training and testing.
The results are reported in \Tref{table:CC_result} and our method outperforms all previous works in every test category.
The training time for the Gehler-Shi set is 24 hours and the test time is 32 minutes (3.4 seconds per image).

Next, we used the Gray-ball set made by Ciurea and Funt~\cite{ciurea2003large}, which contains 11,346 images from 15 video clips of the real world scenes at various places.
Because the images in this dataset are highly correlated within the clips, we have to ensure that the correlated images do not exist both in the training set and the test set. 
For this, we use the 15-fold cross validation which divides the dataset according to 15 video clips, as proposed in~\cite{gijsenij2011computational}. The gray sphere is masked out before the training and the testing. 
We use the linearized version of the Gray-ball set (inverting the gamma correction) and the evaluation is done with recomputed ground truth in \cite{gijsenij2011computational}.
Our method is again evaluated as the most accurate in all categories as shown in \Tref{table:gb_result}.

The third dataset used for the evaluation is the NUS 8-camera set~\cite{cheng2014illuminant} which is composed of images of roughly 200 scenes taken with 8 different DSLR cameras. 
What is interesting about this dataset is that a set of scenes is observed under different cameras and 
the system is trained separately for different cameras in the original work~\cite{cheng2014illuminant, cheng2015effective}.
In contrast, we learn our CNN models by training the data from different cameras altogether to test the learning capability of our system. 
The evaluation for our method is done without the knowledge of the test sample's camera model, and it make the problem more challenging.
Note that the evaluation is done by the 3-fold cross validation using all of images from eight cameras.
\Tref{table:nus_full_result} shows the performance of various methods on this dataset and the results for each camera on the dataset are reported separately as done in~\cite{cheng2015effective}.

On this particular dataset, the performance of our method and the method in \cite{cheng2015effective} are evaluated to be similar. 
Since the dataset contains only 200 scenes, we believe that the accuracy of our method will increase with more data.

\begin{figure*}
\centering
\subfigure{\includegraphics[width=1.0\linewidth]{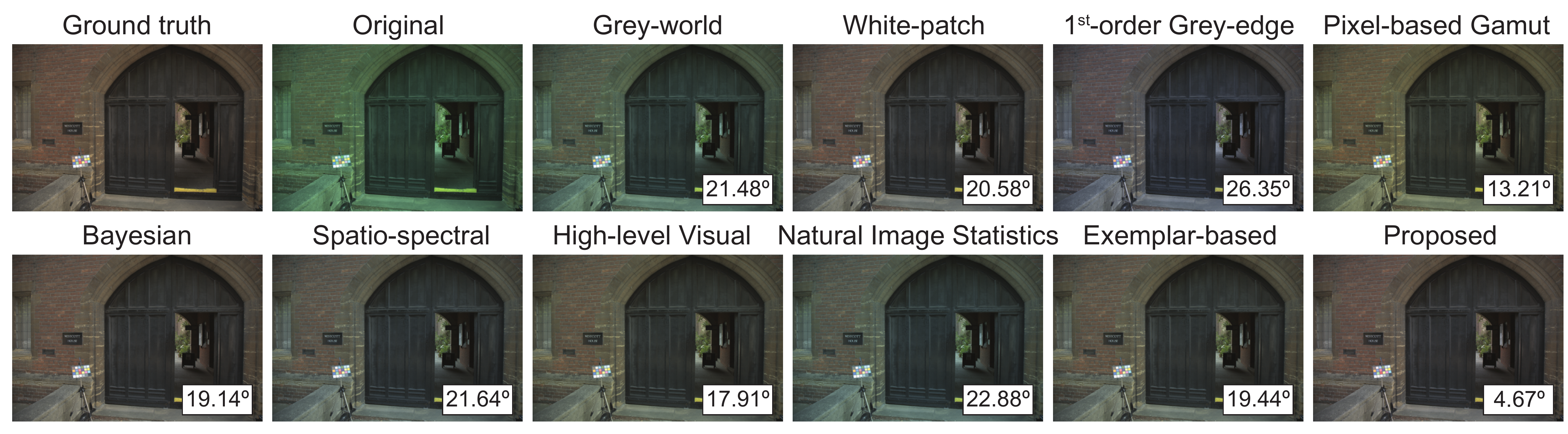}}
\subfigure{\includegraphics[width=1.0\linewidth]{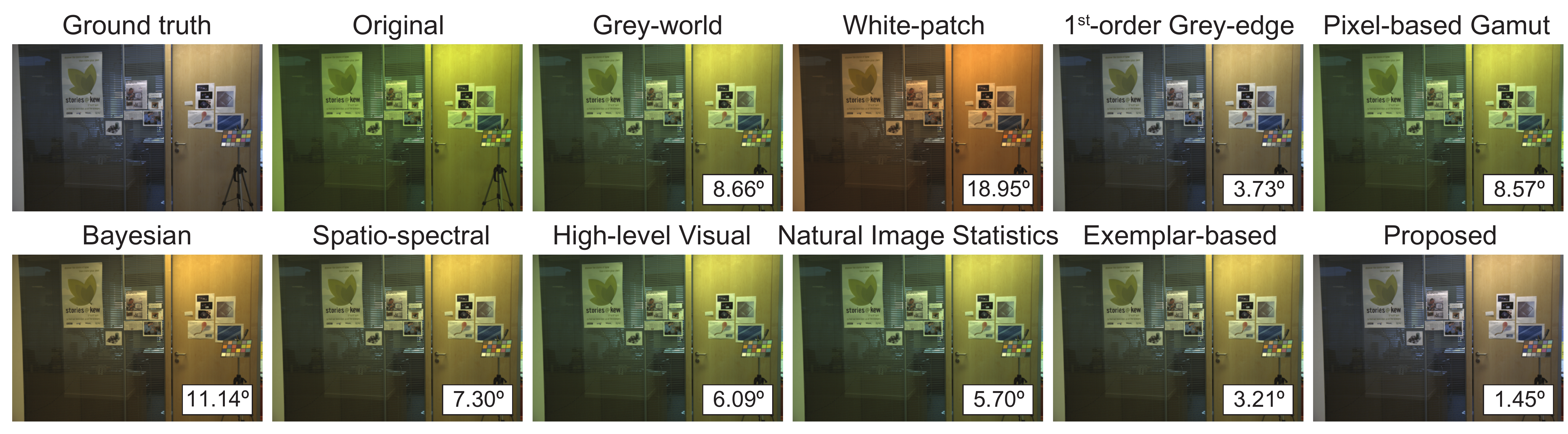}} 
\subfigure{\includegraphics[width=1.0\linewidth]{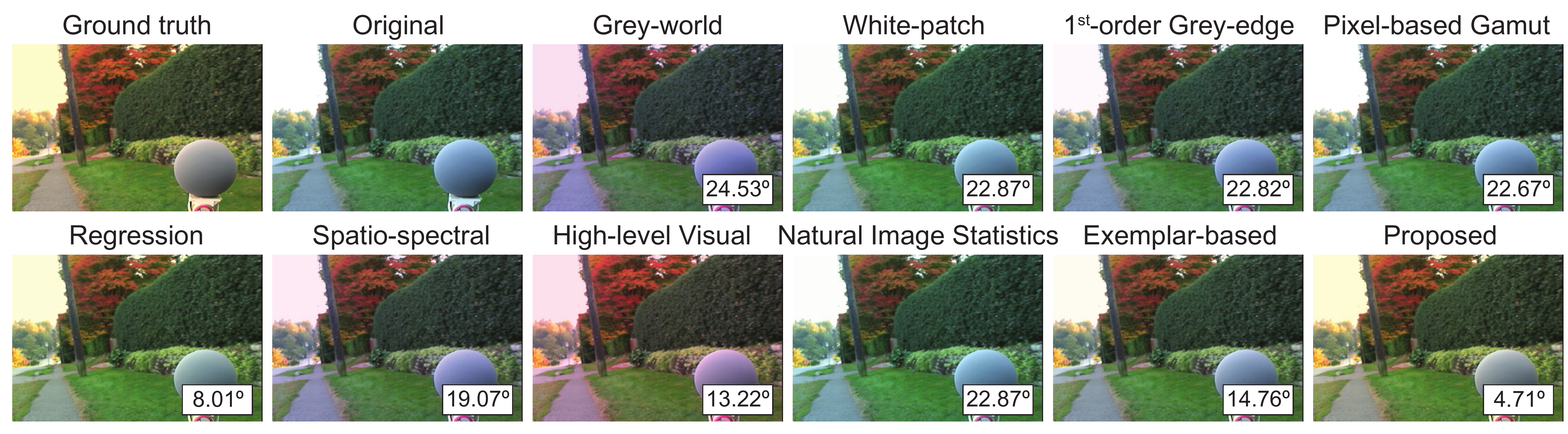}}
\subfigure{\includegraphics[width=1.0\linewidth]{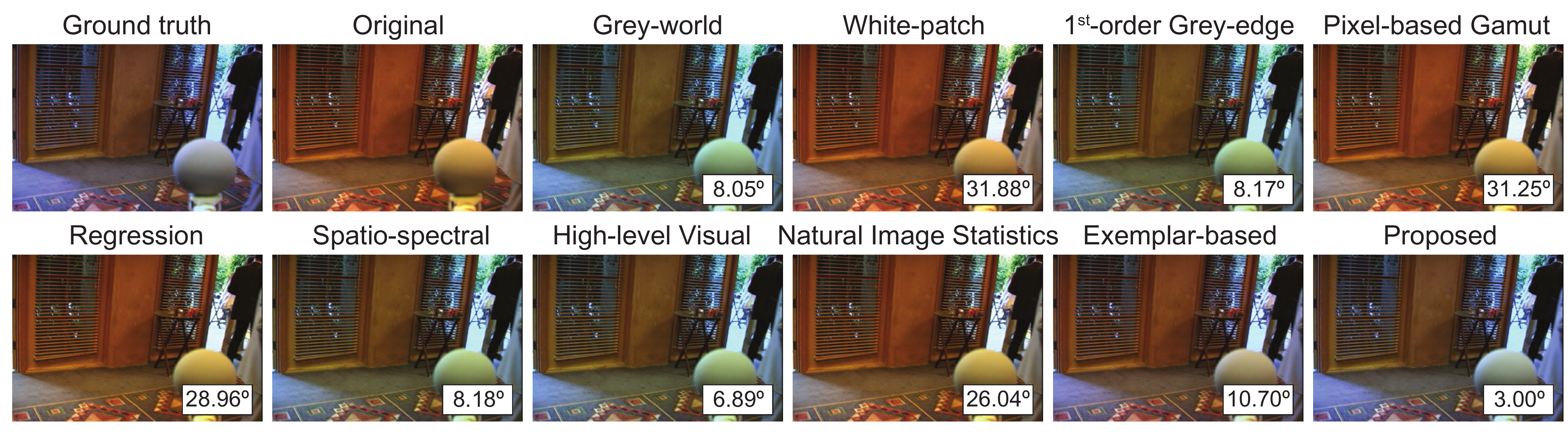}} 
\caption{Some example images corrected using the estimated illuminant from various methods. The angular error is shown on the bottom right corner. Test images are corrected based on following 10 methods: Grey-world~\cite{buchsbaum1980spatial}, White-patch~\cite{land1971lightness}, $1^{st}$-order Grey-edge~\cite{van2007edge}, Pixel-based Gamut~\cite{gijsenij2010generalized}, Bayesian~\cite{gehler2008bayesian}, Spatio-spectral~\cite{chakrabarti2012color}, Natural Image Statistics~\cite{gijsenij2011color}, Exemplar-based~\cite{joze2014exemplar}, and our method. For the existing methods, the results reported in~\cite{cccom} are directly used.} 
\label{fig:qa}
\end{figure*}

In addition to the quantitative evaluations, we provide a qualitative analysis of our method in \fref{fig:qa}.
In \fref{fig:qa}, several images from different datasets are shown along with illumination corrected images using different methods. 
Our method produces visually pleasing outputs on both indoor and outdoor images, even for some extreme cases.

\begin{figure*}
\centering
\subfigure{\includegraphics[width=1.0\linewidth]{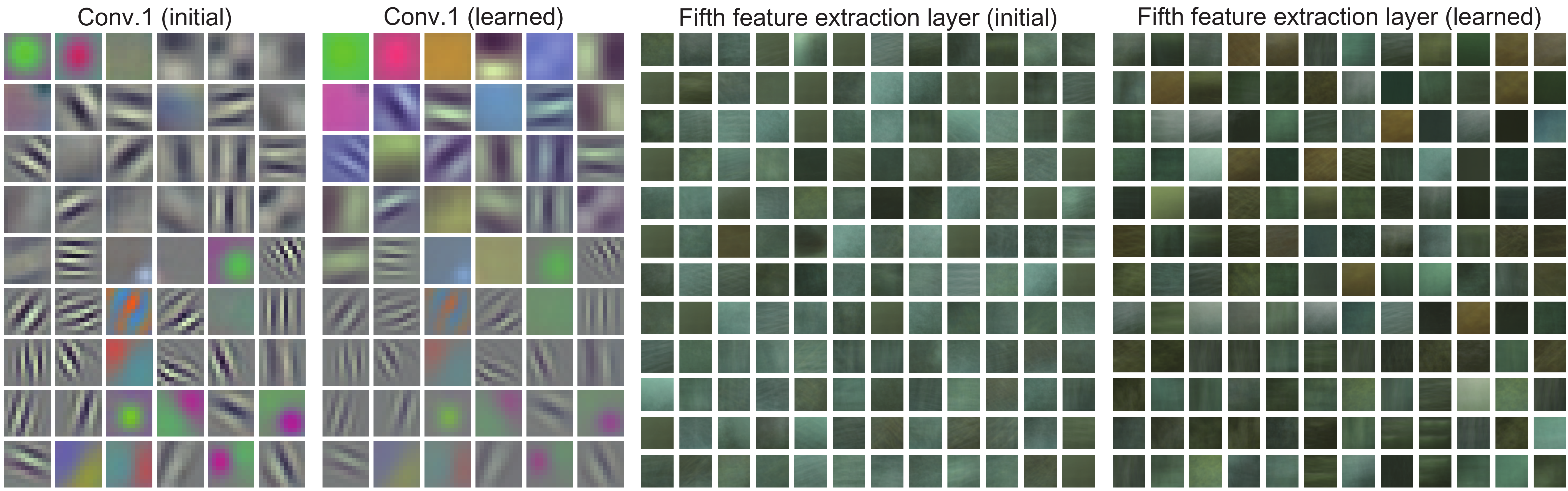}}  
\caption{Visualization of the learned features. First and second columns show first convolution filters before and after the training. Third and fourth columns show visualization of fifth feature extraction layers before and after the training. We can observe that our CNN system becomes sensitive to color information in images after the training.} 
\label{fig:visualize}
\end{figure*}

\begin{figure}
\centering
\subfigure{\includegraphics[width=0.7\linewidth]{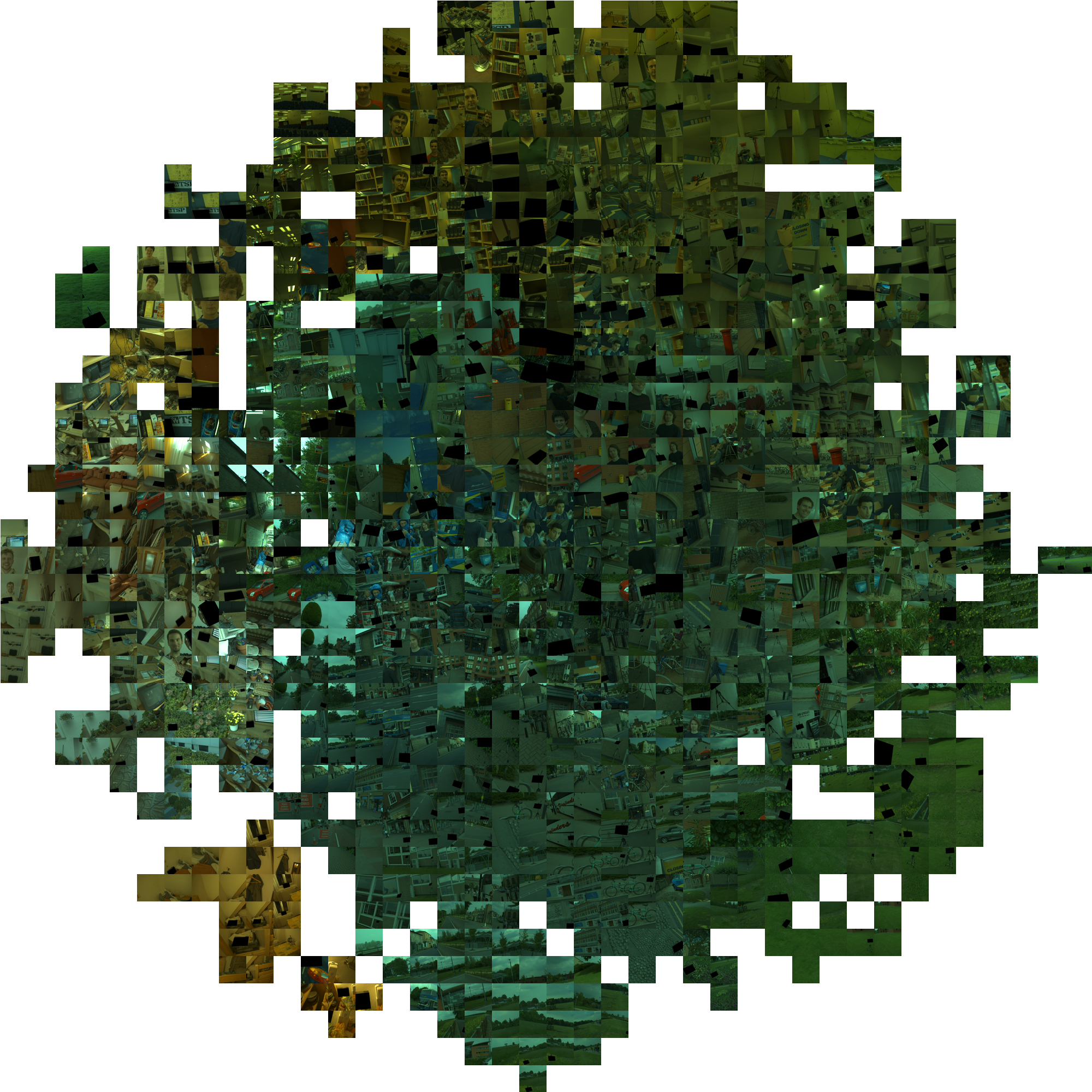}}
\caption{Feature space visualization of te 7th layer of our CNN using the t-SNE method (We recommend to view this figure digitally with zoom). Input patches are located at the resulting two-dimensional feature space. Input patches are well separated according to the scene illuminant.}
\label{fig:embed}
\end{figure}

\section{Discussion}
\label{discussion}
The CNN has shown impressive classification performance in many computer vision problems, especially for recognition problems such as the object recognition and the face recognition.
However, researchers are still trying to understand what is being learned by the CNN and why it works so well~\cite{zeiler2014visualizing, mahendran2015understanding}.

In the previous section, we have shown that using the CNN also improves the computational color constancy performance.
We provide further analysis as an attempt to understand what is being learned by the CNN in our problem.


\subsection{Network Visualization}
To understand what is being learned by the CNN in our problem, we visualized some of the features of the CNN in \fref{fig:visualize}. 
First, we visualize the first layer convolution filters.
The initial filters that are pretrained using the ImageNet database~\cite{deng2009imagenet} and the learned filters after the training with Gehler-Shi set~\cite{Shi_cc} are shown to compare the features for the object classification versus the features for the illumination classification. 
We can observe that chromatic filters which pass specific colors are newly obtained, and some existing achromatic edge filters are replaced with chromatic edge filters.
Next, we averaged top 300 receptive fields that give the largest response at the end of the fifth layer to find out what features are represented by the hidden units as done in \cite{zhou2014learning}. 
Again we compare the results before and after the training, and it is observed that each hidden unit is activated by receptive fields with different color impressions.

Additionally, we visualize the feature space of the 7th layer (4096 dimensional feature for the last fully connected layer) to further analyze what is learned during the training. We use the t-SNE method~\cite{van2008visualizing}, which is one of the famous tools to transform a high-dimensional space into a two-dimensional space (\fref{fig:embed}). We can observe that input patches are well separated according to the scene illuminant in the feature space.

\begin{figure*}
\centering
\subfigure{\includegraphics[width=1.0\linewidth]{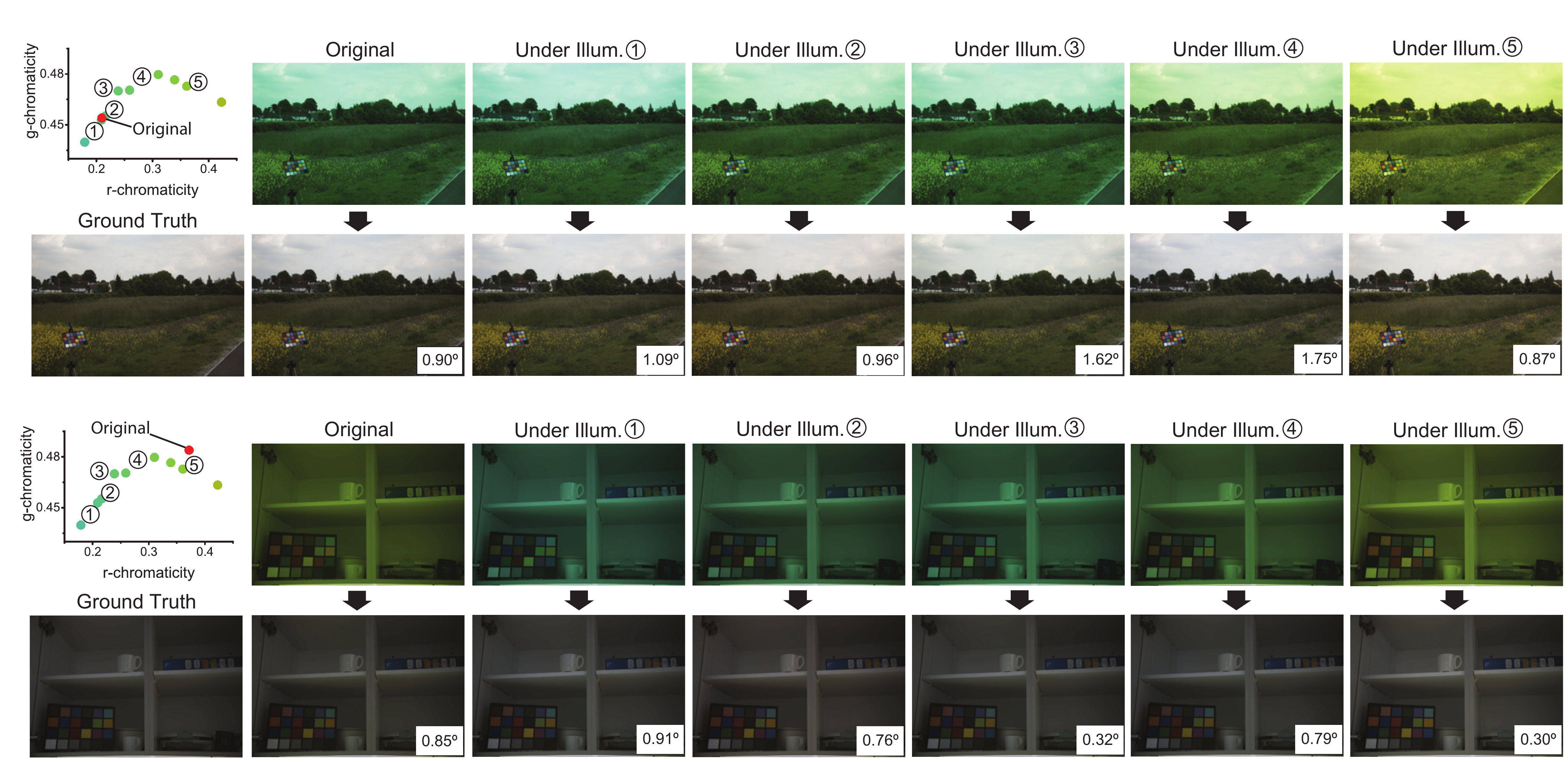}}
\caption{Synthetic experiments for analyzing our CNN system. Synthetic images are created with simulated illumination (the first and the third row). The illuminant colors are estimated by our system and the images are normalized to a canonical illumination (the second and the last row). Since the scene contents are exactly the same, accurate results indicate that the system is learning useful color features.}
\label{fig:synthetic_ex}
\end{figure*}

\subsection{Experiments on Synthetic Images}
As an indirect way to show that our learning system is extracting useful color features, we conducted additional experiments with synthetic images as shown in \fref{fig:synthetic_ex}.
In this experiments, we created synthetic images under different illuminations by multiplying synthetic illumination colors to the original image with the neutral illumination.
We then estimated the illuminant color for each synthetic image using our CNN based estimator, of which the results are shown in \fref{fig:synthetic_ex}.
The accurate results from these experiments indicate that our system is learning useful color features as the scene contents are exactly the same among the images. 

\subsection{Determination and validation of K} \label{diss:K}
The first step of our algorithm is to cluster the illuminants in the training data into $K$ clusters.
There is a trade-off between the difficulty of training the deep network and the final illumination estimation according to the choice of $K$.
With a small $K$, the CNN can be easily trained to classify a test image into the right illumination cluster but the final illumination estimation may fail to recover the accurate illuminant color from the coarse probability distribution.
On the other hand, with a large $K$, the trining becomes more difficult but an accurate illumination color can be computed for correctly classified samples.

\begin{figure}
\centering
\subfigure[Gehler-Shi ($K=25$)]{\includegraphics[width=1.0\linewidth]{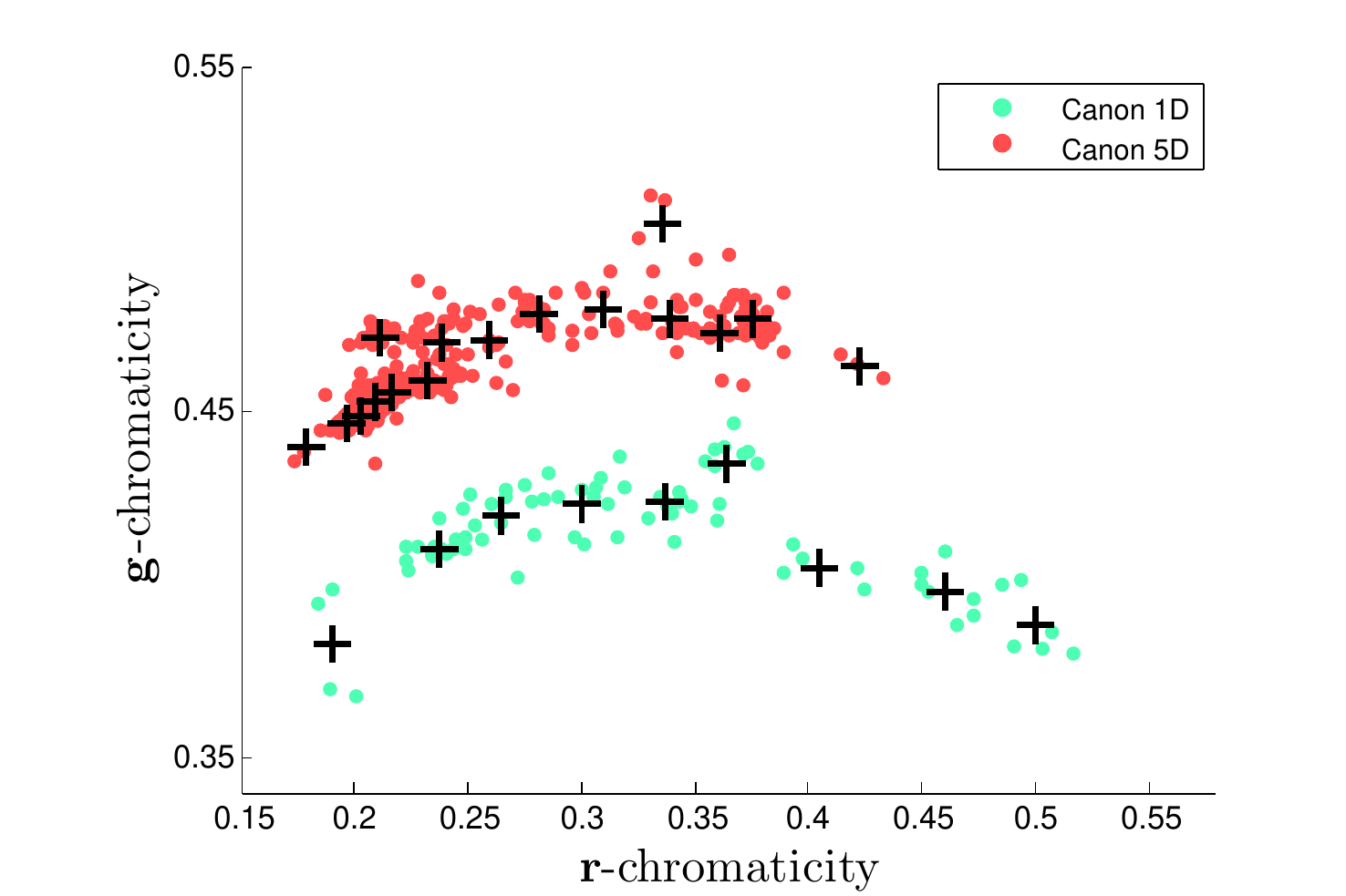}} \\
\subfigure[NUS 8-camera ($K=50$)]{\includegraphics[width=1.0\linewidth]{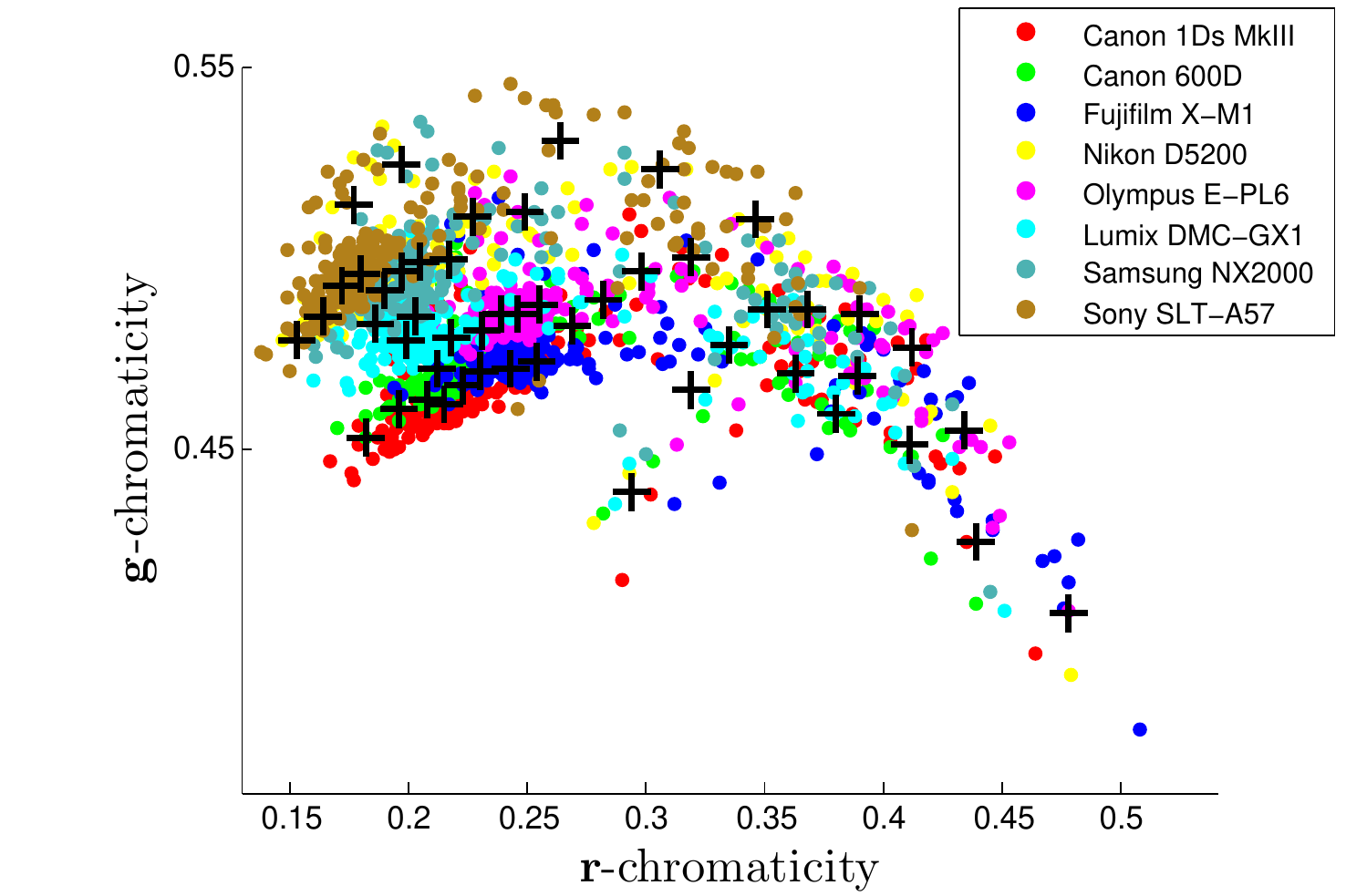}} 
 
\caption{The space of illumination color for two datasets: Gehler-shi~\cite{Shi_cc}, NUS 8-camera~\cite{cheng2014illuminant}. Colored dots are illumination chromaticities from different cameras. Plus marks are cluster centers. The illumination space generally grows with the number of cameras.} 
\label{fig:two_dataset}
\end{figure}

In this paper, we decide the value of $K$ based on the illumination distribution of each dataset, which is closely related to the number of cameras used\footnote{Note that from a practical point of view, testing all possilbe $K$ to find the best option is not viable as the training the deep network requires some time.}.
With a large illumination space, a larger $K$ is used.
The space of the illuminant color grows with the number of cameras as each camera represents the scene with their own color space\footnote{The space of each camera is determined by the spectral response of the color filters of the camera.}.
\Fref{fig:two_dataset} shows the illuminant plots for two datasets captured with two and eight cameras. 
We simply picked the value of $K$ for each dataset based on the number of camera. 
For the SFU Gray-ball set~\cite{ciurea2003large} which was captured using a single camera, we set $K$ to a lower number 20. For the NUS 8-camera set~\cite{cheng2014illuminant} which was captured using eight cameras, we set $K$ to a large number 50.

\begin{figure}
\centering
\subfigure{\includegraphics[width=0.8\linewidth]{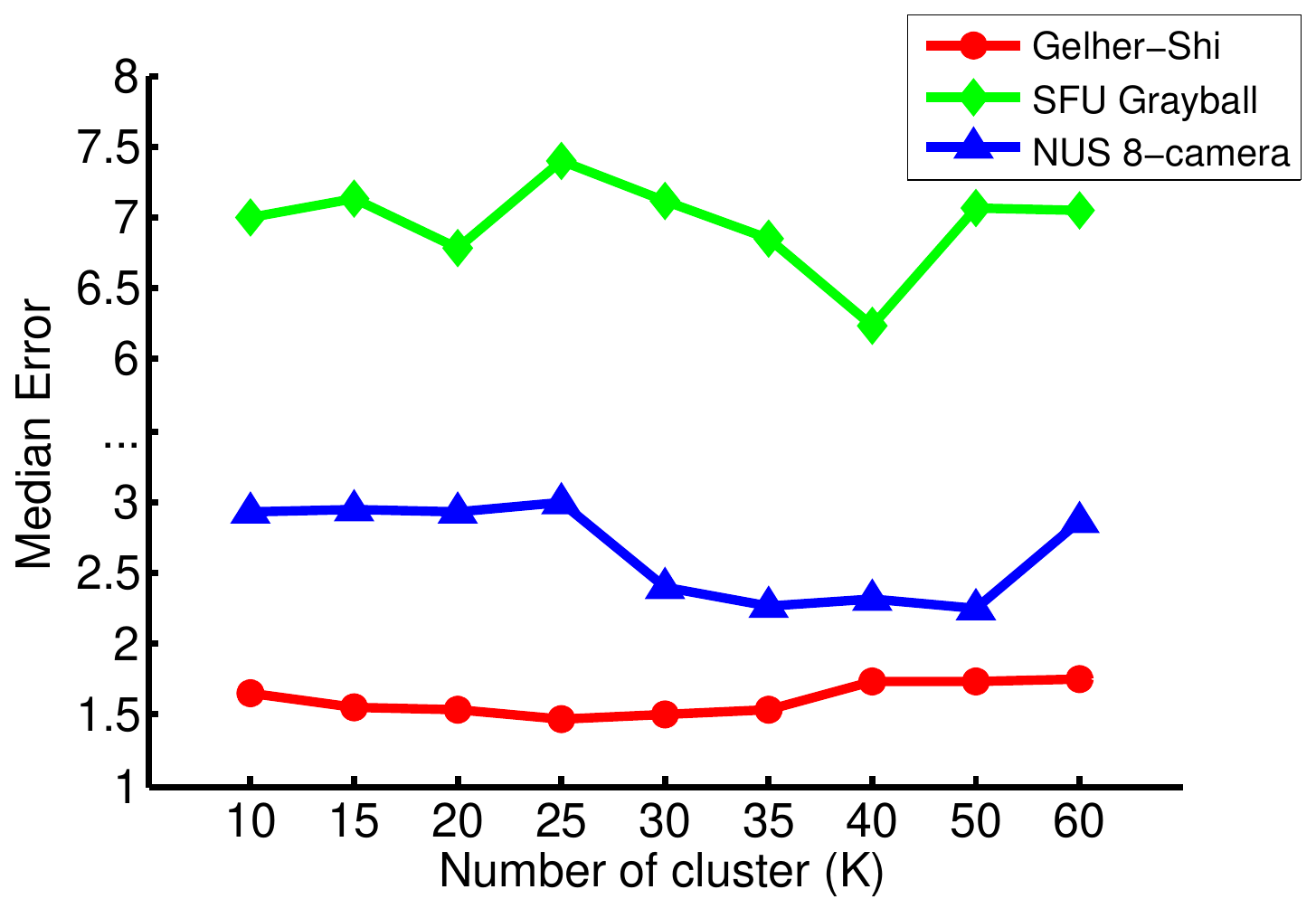}}
\caption{The median angular error with different $K$s. The SFU Gray-ball set~\cite{ciurea2003large} and the NUS 8-camera set~\cite{cheng2014illuminant} requires larger $K$ than the Gehler-Shi set~\cite{Shi_cc}.}
\label{fig:effect_k}
\end{figure}

To validate our method of choosing $K$, we conducted experiments with different $K$'s. 
\Fref{fig:effect_k} shows the median angular error on three dataset with different choice of $K$'s.
Note that the errors reported here are based on a validation set to test the effect of $K$ and they can be different from the results on the full dataset in \Sref{sect:experiments}.
The results of the two datasets (Gehler-Shi~\cite{Shi_cc}, NUS 8-camera~\cite{cheng2014illuminant}) in which the RAW images are provided, are as expected.
However, the SFU Gray-ball set~\cite{ciurea2003large} that provides color-processed images requires more clusters than expected. 
We believe that this is due to the nonlinear in-camera processing~\cite{Kim:2012} (such as tone mapping and gamut mapping), which widens the space of the illumination color. 
However, the selection of $K$ within proper range (10 to 50) shows reasonable performance for all the datasets.

\begin{figure}
\centering
\subfigure[]{\includegraphics[width=0.40\linewidth]{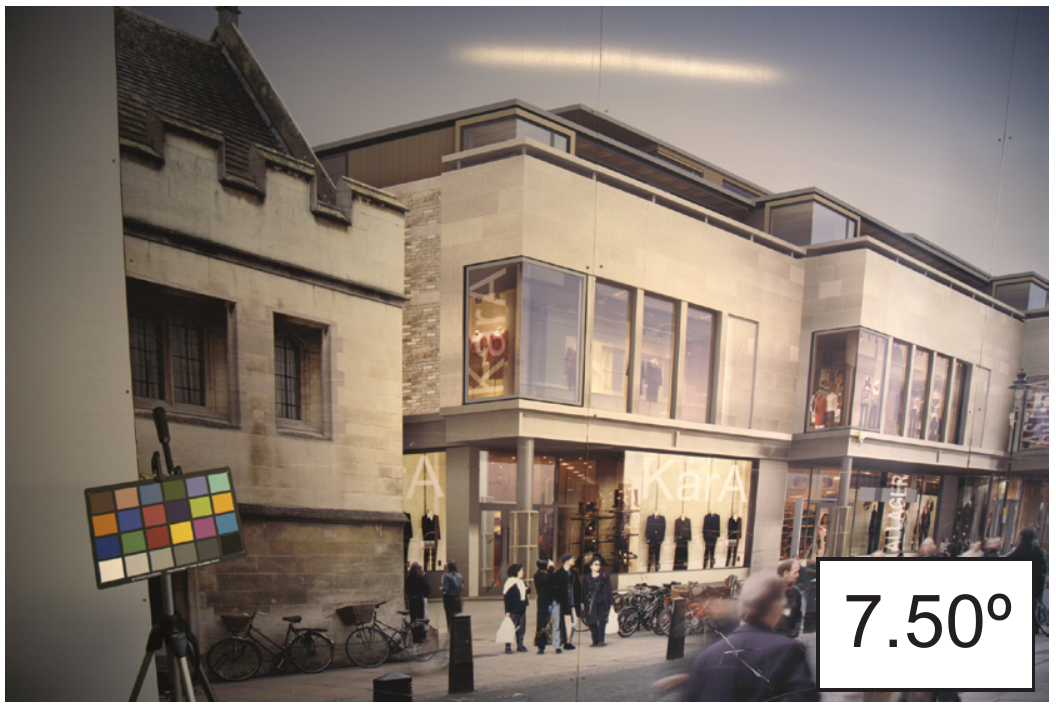}} \nolinebreak
\subfigure[]{\includegraphics[width=0.40\linewidth]{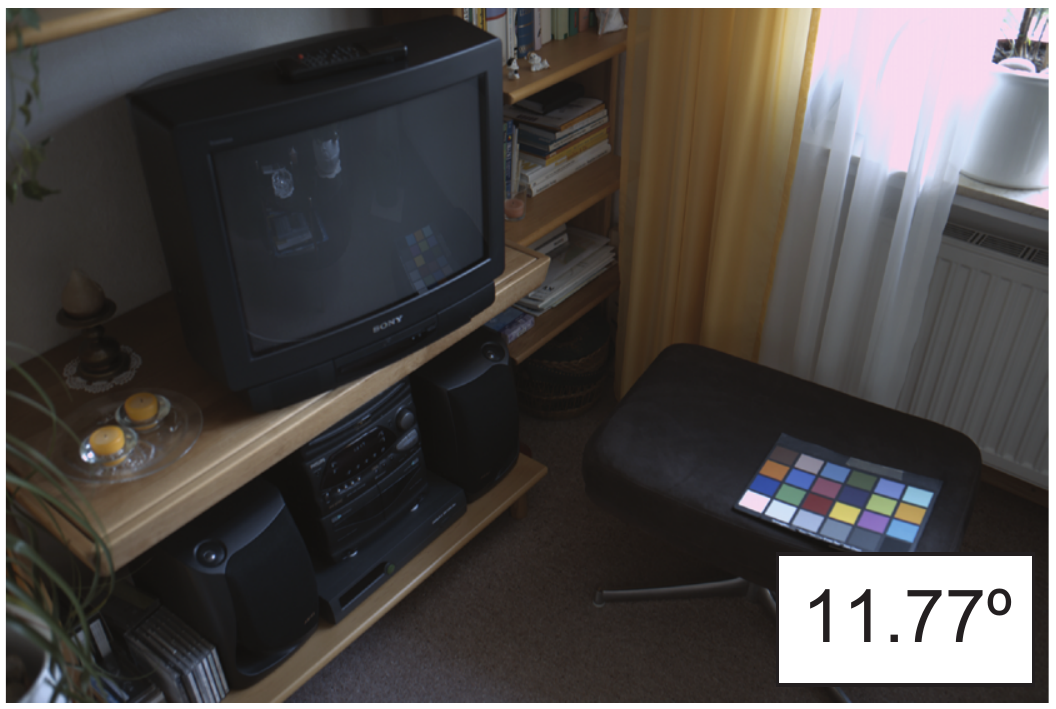}} 
\subfigure[]{\includegraphics[width=0.40\linewidth]{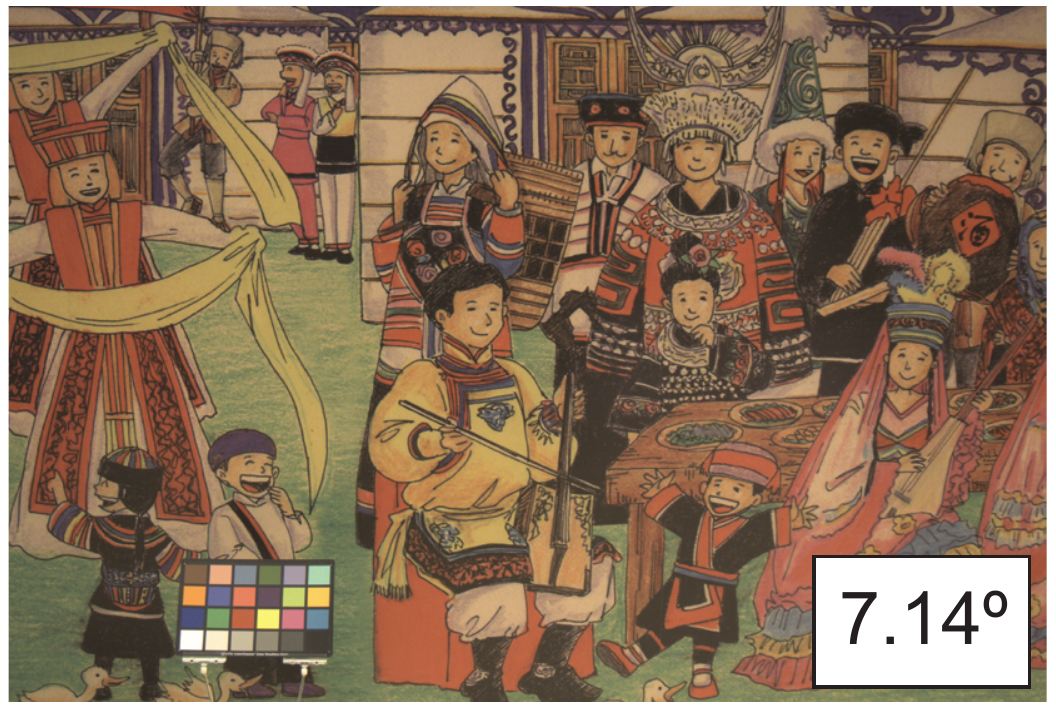}} \nolinebreak
\subfigure[]{\includegraphics[width=0.40\linewidth]{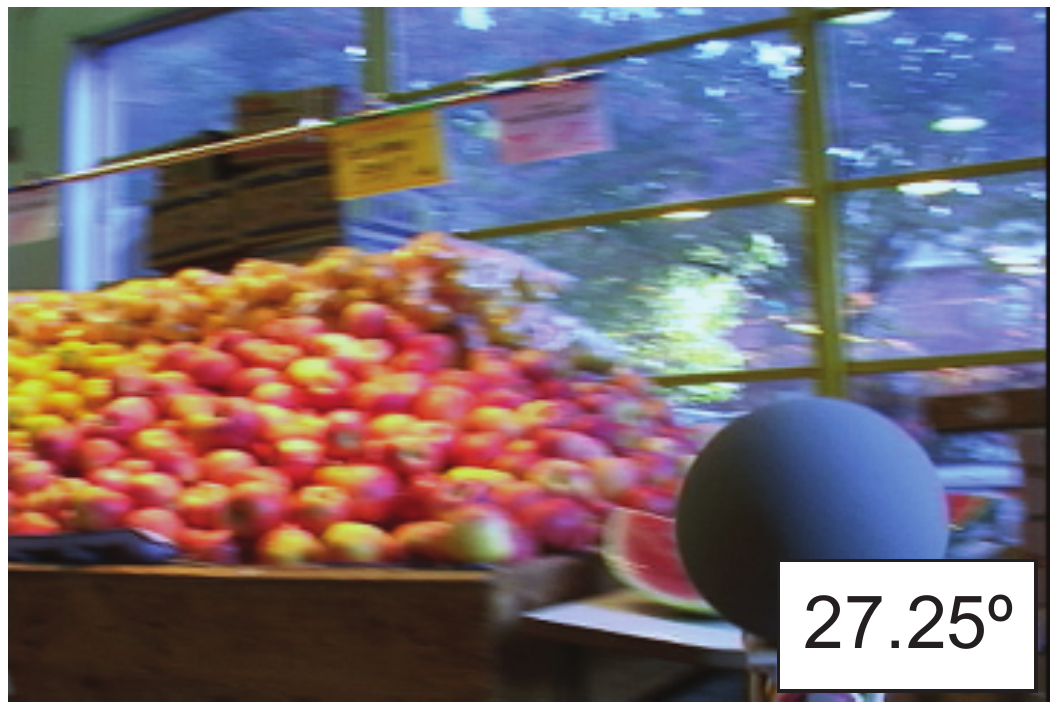}} 
\caption{Some failure cases. Our system fails in the cases of unrealistic scenes((a), (c)), and in case that the ground truth illuminant can not represent the illumination of entire scene((b), (d)).} 
\label{fig:failcase}
\end{figure}

\subsection{Failure Cases}
In \fref{fig:failcase}, we show some failure cases of our work from different datasets.
As can been seen, the system fails when it faces unrealistic scenes that are rarely observed in the training sets such as photos of paintings. 
The system also fails when the assumption of a single illuminant is violated. 

\section{Conclusion}
In this paper, we have introduced a deep learning based computational color constancy algorithm.
The deep learning is making a huge impact in the computer vision research, especially in the high level scene recognition and classification  problems.
The main contribution of this paper is that we have presented a way to apply the CNN for a low-level scene understanding problem in estimating the illuminant color. 
We have validated outstanding performance of our method by comparing it to numerous previous methods with publicly available datasets and
we have also shown some interesting insights to our learning system.
In the future, we are interested in applying different deep architectures to the color constancy problem including the unsupervised methods. 
We are also interested in estimating the illumination on a smaller scale such as on pixel or patch levels, so as to deal with the mixed illumination scenes.
Finally, we believe that having more comprehensive database for color constancy would increase the performance of the deep learning based algorithm even more. 
We plan on putting a lot of effort into designing and implementing a big color constancy database for deep learning.

\section*{References}

\bibliography{mybib}

\end{document}